# Efficient Approximations for the Marginal Likelihood of Incomplete Data Given a Bayesian Network

**David Maxwell Chickering*** **and David Heckerman**
Microsoft Research
Redmond WA 98052-6399
dmax@microsoft.com, heckerma@microsoft.com

## Abstract

We discuss Bayesian methods for learning Bayesian networks when data sets are incomplete. In particular, we examine asymptotic approximations for the marginal likelihood of incomplete data given a Bayesian network. We consider the Laplace approximation and the less accurate but more efficient BIC/MDL approximation. We also consider approximations proposed by Draper (1993) and Cheeseman and Stutz (1995). These approximations are as efficient as BIC/MDL, but their accuracy has not been studied in any depth. We compare the accuracy of these approximations under the assumption that the Laplace approximation is the most accurate. In experiments using synthetic data generated from discrete naive-Bayes models having a hidden root node, we find that (1) the BIC/MDL measure is the least accurate, having a bias in favor of simple models, and (2) the Draper and CS measures are the most accurate.

## 1 Introduction

There is growing interest in methods for learning graphical models from data. We consider Bayesian methods such as those summarized in Heckerman (1995) and Buntine (1996).

A key step in the Bayesian approach to learning graphical models is the computation of the marginal likelihood of a data set given a model. Given a *complete* data set—that is a data set in which each sample contains observations for every variable in the model, the marginal likelihood can be computed exactly and efficiently under certain assumptions (e.g., see Cooper & Herskovits, 1992; Heckerman & Geiger, 1995). In contrast, when observations are missing, including situations where some variables are *hidden* or never observed, the exact determination of the marginal likelihood is typically intractable. Consequently, approximate techniques for computing the marginal likelihood are used. These techniques include Monte Carlo approaches such as Gibbs sampling and importance sampling (Neal, 1993; Madigan & Raftery, 1994), sequential updating methods (Spiegelhalter & Lauritzen, 1990; Cowell, Dawid, & Sebastiani, 1995), and asymptotic approximations (Kass, Tierney, & Kadane, 1988; Kass & Raftery, 1995; Draper, 1993).

In this paper, we examine asymptotic approximations, comparing their accuracy and efficiency. We consider the Laplace approximation (Kass et al., 1988; Kass & Raftery, 1995; Azevedo-Filho & Shachter, 1994) and the Bayesian Information Criterion (BIC) (Schwarz, 1978), which is equivalent to Risannen's (1987) Minimum-Description-Length (MDL) measure. In addition, we consider approximations described by Draper (1993) and Cheeseman and Stutz (1995).

Both theoretical and empirical studies have shown that the Laplace approximation is more accurate than is BIC/MDL (see, e.g., Draper, 1993, and Raftery, 1994). Furthermore, it is well known that the Laplace approximation is significantly less efficient than are the BIC/MDL, Draper, and Cheeseman-Stutz measures. To our knowledge, however, there have been no theoretical or formal empirical studies that compare the accuracy of the Laplace approximation with those of Draper and Cheeseman and Stutz. We describe an experimental comparison of the approximations for learning directed graphical models (Bayesian networks) for discrete variables where one variable is hidden.

*Author's primary affiliation: Computer Science Department, University of California, Los Angeles, CA 90024.

## 2 Background and Motivation

The Bayesian approach for learning Bayesian networks from data is as follows. Given a domain or set of variables $\mathbf{X} = \{X_1, \ldots, X_n\}$, suppose we know that the true joint distribution of $\mathbf{X}$ can be encoded in the Bayesian-network structure $S$. Let $S^h$ denote the hypothesis that this encoding is possible. Also, suppose that we are uncertain about the parameters of the Bayesian network ($\boldsymbol{\theta}_s$, a column vector) that determine the true joint distribution. Given a prior distribution over these parameters and a random sample (data set) $D = \{\mathbf{X}_1 = \mathbf{x}_1, \ldots, \mathbf{X}_N = \mathbf{x}_N\}$ from the true joint distribution, we can apply Bayes' rule to infer the posterior distribution of $\boldsymbol{\theta}_s$:

$$p(\boldsymbol{\theta}_s|D, S^h) = c\ p(D|\boldsymbol{\theta}_s, S^h)\ p(\boldsymbol{\theta}_s|S^h) \qquad (1)$$

where $c$ is a normalization constant. Because $D$ is a random sample, the likelihood $p(D|\boldsymbol{\theta}_s, S^h)$ is simply the product of the individual likelihoods

$$p(D|\boldsymbol{\theta}_s, S^h) = \prod_{l=1}^{N} p(\mathbf{x}_l|\boldsymbol{\theta}_s, S^h)$$

Given some quantity of interest that is a function of the network-structure hypothesis and its parameters, $f(S^h, \boldsymbol{\theta}_s)$, we can compute its expectation, given $D$, as follows:

$$E(f(S^h, \boldsymbol{\theta}_s)|D, S^h) = \int f(S^h, \boldsymbol{\theta}_s)\ p(\boldsymbol{\theta}_s|D, S^h)\ d\boldsymbol{\theta}_s \qquad (2)$$

Consider the case where the variables $\mathbf{X}$ are discrete. Let $\mathbf{Pa}_i$ denote the set of variables corresponding to the parents of $X_i$. Let $x_i^k$ and $\mathbf{pa}_i^j$ denote the $k$th possible state of $X_i$ and the $j$th possible state of $\mathbf{Pa}_i$, respectively. Also, let $r_i$ and $q_i$ denote the number of possible states of $X_i$ and $\mathbf{Pa}_i$, respectively. Assuming that there are no logical constraints on the true joint probabilities other than those imposed by the network structure $S$, the parameters $\boldsymbol{\theta}_s$ correspond to the true probabilities (i.e., long-run fractions) associated with the Bayesian-network structure. In particular, $\boldsymbol{\theta}_s$ is the set of parameters $\theta_{ijk}$ for all possible values of $i, j$, and $k$, where $\theta_{ijk}$ is the true probability that $X_i = x_i^k$ given $\mathbf{Pa}_i = \mathbf{pa}_i^j$. We use the notation

$$\boldsymbol{\theta}_{ij} = (\theta_{ijk})_{k=1}^{r_i} \qquad \boldsymbol{\theta}_i = (\boldsymbol{\theta}_{ij})_{j=1}^{q_i} \qquad \boldsymbol{\theta}_s = (\boldsymbol{\theta}_i)_{i=1}^{n}$$

The likelihood for a random sample with no missing observations is given by

$$p(D|\boldsymbol{\theta}_s, S^h) = \prod_{i=1}^{n}\prod_{j=1}^{q_i}\prod_{k=1}^{r_i} \theta_{ijk}^{N_{ijk}}$$

where $N_{ijk}$ are the sufficient statistics for the likelihood—the number of samples in $D$ in which $X_i = x_i^k$ and $\mathbf{Pa}_i = \mathbf{pa}_i^j$. Consequently, we can compute the posterior distribution of $\boldsymbol{\theta}_s$ using Equation 1. This computation is especially simple when (1) the parameter sets $\boldsymbol{\theta}_{ij}$ are mutually independent—an assumption we call *parameter independence*—and (2) the prior distribution for each parameter set $\boldsymbol{\theta}_{ij}$ is a Dirichlet distribution

$$p(\boldsymbol{\theta}_{ij}|S^h) = c \prod_{k=1}^{r_i} \theta_{ijk}^{\alpha_{ijk}-1} \qquad (3)$$

where $c$ is a normalization constant and the $\alpha_{ijk} > 0$ may depend on the network structure $S$.

Making the problem more difficult, suppose that we are also uncertain about which structure encodes the true distribution. Given a prior distribution over the possible network-structure hypotheses, we can compute the corresponding posterior distribution using Bayes rule:

$$\begin{aligned} p(S^h|D) &= c\ p(S^h)\ p(D|S^h) \qquad (4)\\ &= c\ p(S^h) \int p(D|\boldsymbol{\theta}_s, S^h)\ p(\boldsymbol{\theta}_s|S^h)\ d\boldsymbol{\theta}_s \end{aligned}$$

Given some quantity of interest, $f(S^h, \boldsymbol{\theta}_s)$, we can compute its expectation, given $D$:

$$E(f(S^h, \boldsymbol{\theta}_s)|D) = \sum_{S^h} p(S^h|D) \int f(S^h, \boldsymbol{\theta}_s)\ p(\boldsymbol{\theta}_s|D, S^h)\ d\boldsymbol{\theta}_s$$

This Bayesian approach is an example of what statisticians call *model averaging.* The key computation here is that of $p(D|S^h)$, known as the *marginal likelihood of $D$ given $S$*, or simply the marginal likelihood of $S$. In the remainder of the paper, we assume that priors over network structure are uniform, so that relative posterior probability and marginal likelihood are the same.

When we can not use prior knowledge to restrict the set of possible network structures to a manageable number, we can select (typically) one model $S$ and use Equation 2 to approximate the true expectation of $f(S^h, \boldsymbol{\theta}_s)$. This approximation is an example of *model selection.* In practice, one selects a model by using some search procedure that produces candidate network structures, applying a *scoring function* to each found structure, and retaining the structure with the highest score. One reasonable scoring function is the log marginal likelihood: $\log p(D|S^h)$.

Dawid (1984) notes the following interesting interpretation of the log marginal likelihood as a scoring function. Suppose we use a model $S$ to predict the probability of each sample in $D$ given the previously observed samples in $D$, and assign a utility to each pre-diction using the log proper scoring rule. The utility for the first prediction will be $\log p(\mathbf{x}_1|S^h)$. We make this prediction based solely on the prior distribution for $\boldsymbol{\theta}_s$. The utility for the second prediction will be $\log p(\mathbf{x}_2|\mathbf{x}_1, S^h)$. We compute this prediction by training the network structure using only the first sample in $D$. The utility for the $i$th prediction will be $\log p(\mathbf{x}_i|\mathbf{x}_1, \ldots, \mathbf{x}_{i-1}, S^h)$. We make this prediction by training the network structure with the first $i-1$ samples in $D$. Summing these utilities (as one does with this proper scoring rule), we obtain

$$\sum_{i=1}^{N} \log p(\mathbf{x}_i|\mathbf{x}_1, \ldots, \mathbf{x}_{i-1}, S^h)$$

By Bayes' rule this sum is equal to the log marginal likelihood $\log p(D|S^h)$, and is independent of the order in which we process the samples in $D$. Thus, the network structure with the highest log marginal likelihood is precisely the model that is the best sequential predictor of the data $D$ according to the log scoring rule.

When the random sample $D$ is complete, parameters are independent, and parameter priors are Dirichlet, the computation of the marginal likelihood is straightforward:

$$p(D|S^h) = \prod_{i=1}^{n} \prod_{j=1}^{q_i} \frac{\Gamma(\alpha_{ij})}{\Gamma(\alpha_{ij} + N_{ij})} \cdot \prod_{k=1}^{r_i} \frac{\Gamma(\alpha_{ijk} + N_{ijk})}{\Gamma(\alpha_{ijk})} \quad (5)$$

This formula was first derived by Cooper and Herskovits (1992). Heckerman et al. (1995) refer to this formula in conjunction with the structure prior as the *B*ayesian *D*irichlet (BD) scoring function.

When the random sample $D$ is incomplete, the exact computation of the marginal likelihood is intractable for real-world problems (e.g., see Cooper & Herskovits, 1992). Thus, approximations are required. In this paper, we consider asymptotic approximations.

One well-known asymptotic approximation is the *Laplace* or *Gaussian* approximation (Kass et al., 1988; Kass & Raftery, 1995; Azevedo-Filho & Shachter, 1994). The idea behind the Laplace approximation is that, for large amounts of data, $p(\boldsymbol{\theta}_s|D, S^h) \propto p(D|\boldsymbol{\theta}_s, S^h) \cdot p(\boldsymbol{\theta}_s|S^h)$ can often be approximated as a multivariate Gaussian distribution. Consequently,

$$p(D|S^h) = \int p(D|\boldsymbol{\theta}_s, S^h)\ p(\boldsymbol{\theta}_s|S^h)\ d\boldsymbol{\theta}_s \quad (6)$$

can be evaluated in closed form. In particular, let

$$g(\boldsymbol{\theta}_s) \equiv \log(p(D|\boldsymbol{\theta}_s, S^h) \cdot p(\boldsymbol{\theta}_s|S^h))$$

Let $\tilde{\boldsymbol{\theta}}_s$ be the (vector) value of $\boldsymbol{\theta}_s$ for which the posterior probability of $\boldsymbol{\theta}_s$ is a maximum:

$$\tilde{\boldsymbol{\theta}}_s = \arg\max_{\boldsymbol{\theta}_s} \{p(\boldsymbol{\theta}_s|D, S^h)\} = \arg\max_{\boldsymbol{\theta}_s} \{g(\boldsymbol{\theta}_s)\}$$

The quantity $\tilde{\boldsymbol{\theta}}_s$ is known as the maximum *a posteriori* probability (MAP) value of $\boldsymbol{\theta}_s$. Expanding $g(\boldsymbol{\theta}_s)$ about $\tilde{\boldsymbol{\theta}}_s$, we obtain

$$g(\boldsymbol{\theta}_s) \approx g(\tilde{\boldsymbol{\theta}}_s) + -\frac{1}{2}(\boldsymbol{\theta}_s - \tilde{\boldsymbol{\theta}}_s)^t A (\boldsymbol{\theta}_s - \tilde{\boldsymbol{\theta}}_s) \quad (7)$$

where $(\boldsymbol{\theta}_s - \tilde{\boldsymbol{\theta}}_s)^t$ is the transpose of column vector $(\boldsymbol{\theta}_s - \tilde{\boldsymbol{\theta}}_s)$, and $A$ is the negative Hessian of $g(\boldsymbol{\theta}_s)$ evaluated at $\tilde{\boldsymbol{\theta}}_s$. Substituting Equation 7 into Equation 6, integrating, and taking the logarithm of the result, we obtain the Laplace approximation:

$$\log p(D|S^h) \approx \log p(D|\tilde{\boldsymbol{\theta}}_s, S^h) + \log p(\tilde{\boldsymbol{\theta}}_s|S^h) + \frac{d}{2}\log(2\pi) - \frac{1}{2}\log|A| \quad (8)$$

where $d$ is the dimension of the model $S$ given $D$ in the region of $\tilde{\boldsymbol{\theta}}_s$. For a Bayesian network with discrete variables, this dimension is typically the number of parameters of the network structure, $\sum_{i=1}^{n} q_i(r_i - 1)$. (When enough data are missing—for example, when one or more variables are hidden—it may be that the dimension is lower. See Geiger et al. in this proceedings for a discussion.) Kass et al. (1988) have shown that, under certain regularity conditions, errors in this approximation are bounded by $O(1/N)$, where $N$ is the number of samples in $D$.

A more efficient but less accurate approximation is obtained by retaining only those terms in Equation 8 that increase with $N$: $\log p(D|\tilde{\boldsymbol{\theta}}_s, S^h)$, which increases linearly with $N$, and $\log|A|$, which increases as $d \log N$. Also, for large $N$, $\tilde{\boldsymbol{\theta}}_s$ can be approximated by the *maximum likelihood* (ML) value of $\boldsymbol{\theta}_s$, $\hat{\boldsymbol{\theta}}_s$, the vector value of $\boldsymbol{\theta}_s$ for which $p(D|\boldsymbol{\theta}_s, S^h)$ is a maximum. Thus, we obtain

$$\log p(D|S^h) \approx \log p(D|\hat{\boldsymbol{\theta}}_s, S^h) - \frac{d}{2}\log N \quad (9)$$

This approximation is called the *Bayesian information criterion* (BIC), and was first derived by Schwarz (1978).

Given regularity conditions similar to those for the Laplace approximation, BIC is accurate to $O(1)$. That is, for large $N$, the error bounds of the approximation do not increase as $N$ increases.[1] Thus, if we use BIC to select one of a set of models, we will select a model whose posterior probability is a maximum, when $N$ becomes sufficiently large. We say that BIC is *asymptotically correct*. By this definition, the Laplace approximation is also asymptotically correct.

[1] Under some conditions, the BIC is accurate to $O(N^{-1/2})$ (Kass & Wasserman, 1996). These conditions do not apply to the models we examine in our experiments.

The BIC approximation is interesting in several respects. First, it does not depend on the prior. Consequently, we can use the approximation without assessing a prior.[2] Second, the approximation is quite intuitive. Namely, it contains a term measuring how well the model with parameters set to an ML value predicts the data ($\log p(D|\hat{\boldsymbol{\theta}}_s, S^h)$) and a term that punishes the complexity of the model ($d/2 \; \log N$). Third, the BIC approximation is exactly the additive inverse of the Minimum Description Length (MDL) scoring function described by Rissanen (1987).

Draper (1993) suggests another approximation to Equation 8, in which the term $\frac{d}{2}\log(2\pi)$ is retained:

$$\log p(D|S^h) \approx \log p(D|\hat{\boldsymbol{\theta}}_s, S^h) - \frac{d}{2}\log N + \frac{d}{2}\log(2\pi) \tag{10}$$

This measure is asymptotically correct under the same conditions as those for BIC/MDL. For finite data sets, however, Draper (1993) mentions that he has found his approximation to be better than BIC/MDL. We shall refer to Equation 10 as the *Draper* scoring function.

To compute the Laplace approximation, we must compute the negative Hessian of $g(\boldsymbol{\theta}_s)$ evaluated at $\tilde{\boldsymbol{\theta}}_s$. Meng and Rubin (1991) describe a numerical technique for computing the second derivatives. Raftery (1995) shows how to approximate the Hessian using likelihood-ratio tests that are available in many statistical packages. Thiesson (1995) demonstrates that, for discrete variables, the second derivatives can be computed using Bayesian-network inference.

When computing any of these approximations, we must determine $\tilde{\boldsymbol{\theta}}_s$ or $\hat{\boldsymbol{\theta}}_s$. One technique for finding a maximum is gradient ascent, where we follow the derivatives of $g(\boldsymbol{\theta}_s)$ or the likelihood to a local maximum. Buntine (1994), Russell et al. (1995), and Thiesson (1995) discuss how to compute derivatives of the likelihood for a Bayesian network with discrete variables.

A more efficient technique for identifying a local MAP or ML value of $\boldsymbol{\theta}_s$ is the EM algorithm (Dempster, Laird, & Rubin, 1977). Applied to Bayesian networks for discrete variables, the EM algorithm works as follows. First, we assign values to $\boldsymbol{\theta}_s$ somehow (e.g., at random). Next, we compute the *expected sufficient statistics* for the missing entries in the data:

$$E(N_{ijk}|\boldsymbol{\theta}_s, S^h) = \sum_{l=1}^{N} p(x_i^k, \mathbf{pa}_i^j|\mathbf{x}_l, \boldsymbol{\theta}_s, S^h) \tag{11}$$

When $X_i$ and all the variables in $\mathbf{Pa}_i$ are observed in sample $\mathbf{x}_l$, the term for this sample requires a trivial computation: it is either zero or one. Otherwise, we can use any Bayesian network inference algorithm to evaluate the term. This computation is called the *E step* of the EM algorithm.

Next, we use the expected sufficient statistics as if they were actual sufficient statistics from a complete random sample $D'$. If we are doing a MAP calculation, we compute the values of $\boldsymbol{\theta}_s$ that maximize $p(\boldsymbol{\theta}_s|D', S^h)$:

$$\theta_{ijk} = \frac{E(N_{ijk}|\boldsymbol{\theta}_s) + \alpha_{ijk} - 1}{E(N_{ij}|\boldsymbol{\theta}_s) + \alpha_{ij} - r_i}$$

If we are doing an ML calculation, we compute the values of $\boldsymbol{\theta}_s$ that maximize $p(D'|\boldsymbol{\theta}_s, S^h)$:

$$\theta_{ijk} = \frac{E(N_{ijk}|\boldsymbol{\theta}_s)}{E(N_{ij}|\boldsymbol{\theta}_s)}$$

This assignment is called the *M step* of the EM algorithm. Dempster et al. (1977) showed that, under certain regularity conditions, iteration of the E and M steps will converge to a local maximum. The EM algorithm assumes parameter independence,[3] and is typically used whenever the expected sufficient statistics can be computed efficiently (e.g., discrete, Gaussian, and Gaussian-mixture distributions).

In the EM algorithm, we treat expected sufficient statistics as if they are actual sufficient statistics. This use suggests another approximation for the marginal likelihood:

$$\log p(D|S^h) \approx \log p(D'|S^h) \tag{12}$$

where $D'$ is an imaginary data set that is consistent with the expected sufficient statistics computed using an E step at a local ML value for $\boldsymbol{\theta}_s$. For discrete variables, this approximation is given by the logarithm of the right-hand-side of Equation 5, where $N_{ijk}$ is replaced by $E(N_{ijk}|\hat{\boldsymbol{\theta}}_s)$. We call this scoring function the *marginal likelihood of the expected data* or MLED.

MLED has two desirable properties. One, because it computes a marginal likelihood, it punishes model complexity as does the Laplace, Draper, and BIC/MDL measures. Two, because $D'$ is a complete (albeit imaginary) data set, the computation of the measure is efficient.

[2] One of the technical assumptions used to derive this approximation is that the prior be non-zero around $\hat{\boldsymbol{\theta}}_s$.

[3] Actually, some parameter sets may be equal, provided these sets are mutually independent.

One problem with this scoring function is that it may not be asymptotically correct. In particular, assuming the BIC/MDL regularity conditions apply, we have

$$\log p(D'|S^h) = \log p(D'|\hat{\boldsymbol{\theta}}_s, S^h) \; - \frac{d'}{2}\log N + O(1)$$

where $d'$ is the dimension of the model $S$ given data $D'$ in the region around $\hat{\boldsymbol{\theta}}_s$—that is, the number of parameters of $S$. As $N$ increases, the difference between $p(D|\hat{\boldsymbol{\theta}}_s, S^h)$ and $p(D'|\hat{\boldsymbol{\theta}}_s, S^h)$ may increase. Also, as we have discussed, it may be that $d' > d$. In either case, MLED will not be asymptotically correct. A simple modification to MLED addresses these problems:

$$\begin{aligned} \log p(D|S^h) \quad \approx \quad & \log p(D'|S^h) \\ & - \log p(D'|\hat{\boldsymbol{\theta}}_s, S^h) + \frac{d'}{2}\log N \\ & + \log p(D|\hat{\boldsymbol{\theta}}_s, S^h) - \frac{d}{2}\log N \end{aligned} \tag{13}$$

Equation 12 (without the correction to dimension) was first proposed by Cheeseman and Stutz (1995) as a scoring function for AutoClass, an algorithm for data clustering. We shall refer to Equation 13 as the *Cheeseman-Stutz* (CS) scoring function. We note that both the MLED and CS scoring functions can easily be extended to the directed Gaussian-mixture models described in Lauritzen and Wermuth (1989) and to undirected Gaussian-mixture models.

The accuracy of these approximations, which we examine in the following two sections, must be balanced against their computation costs. The evaluation of CS, MLED, Draper, and BIC/MDL is dominated by the determination of the MAP or ML. The time complexity of this task is $O(edNi)$, where $e$ is the number of EM iterations and $i$ is the cost of inference in Equation 11. The evaluation of the Laplace approximation is dominated by the computation of the determinant of the negative Hessian $A$. The time complexity of this computation (using Thiesson's 1995 method) is $O(d^2Ni + d^3)$. Typically $e < d$ and $d < N$ so that the Laplace approximation is the least efficient, having complexity $O(d^2Ni)$.

## 3 Experimental Design

As mentioned, the Laplace approximation is known to be more accurate than the BIC/MDL and Draper approximations. In contrast, to our knowledge, no theoretical (or empirical) work has been done comparing the Laplace approximation with the CS or MLED approximations. Nonetheless, in our experiments, we *assume* that the Laplace approximation is the most accurate of the approximations, and measure the accuracy of the other approximations using the Laplace approximation as a gold standard. We can not verify our assumption, because exact computations of the marginal likelihood are not possible for the models that we consider. Thus, the results of our experiments must be interpreted with caution. In particular, we can not rule out the possibility that the CS or MLED approximations are better than the Laplace approximation.

We evaluated the accuracy of the CS, MLED, Draper, and BIC/MDL approximations relative to that of the Laplace approximation using synthetic models containing a single hidden variable. For reasons discussed in Section 4, we limited our synthetic networks to naive-Bayes models for discrete variables (also known as discrete mixture models). A naive-Bayes model for variables $\{C, X_1, \ldots, X_n\}$ encodes the assertion that $X_1, \ldots, X_n$ are mutually independent, given $C$. The network structure for this model contains the single root node $C$ and leaf nodes $X_i$ each having only $C$ as a parent. (We use the same notation to refer to a variable and its corresponding node in the network structure.) We generated a variety of naive-Bayes models by varying the number of states of $C$ ($c$) and the number of observed variables $n$ (all of which are binary). We determined the parameters of each model by sampling from the uniform (Dirichlet) distribution ($\alpha_{ijk} = 1$).

We sampled data from a model so as to make the root node $C$ a hidden variable. Namely, we sampled data from a model using the usual Monte-Carlo approach where we first sampled a state $C = c$ according to $p(C)$ and then sampled a state of each $X_i$ according to $p(X_i|C = c)$. We then discarded the samples of $C$, retaining only the samples of $X_1, \ldots, X_n$.

In a single experiment, we first generated a model for a given $n$ and $c$, and subsequently five data sets for a given sample size $N$. Next, we approximated the marginal likelihood for each data set given a series of *test models* that were identical to the synthesized model, except we allowed the number of states of the hidden variable to vary. Finally, we compared the different approximations of the marginal likelihood in the context of both model averaging and model selection. To compare the approximations for model averaging, we simply compared plots of log marginal likelihood versus states of the hidden variable in the test model directly. To compare the approximations for model selection, we compared the number of states of the hidden variable selected using a given approximation with the number of states selected using the Laplace approximation.

We initialized the EM algorithm as follows. First, we initialized 64 copies of the parameters $\boldsymbol{\theta}_s$ at random,

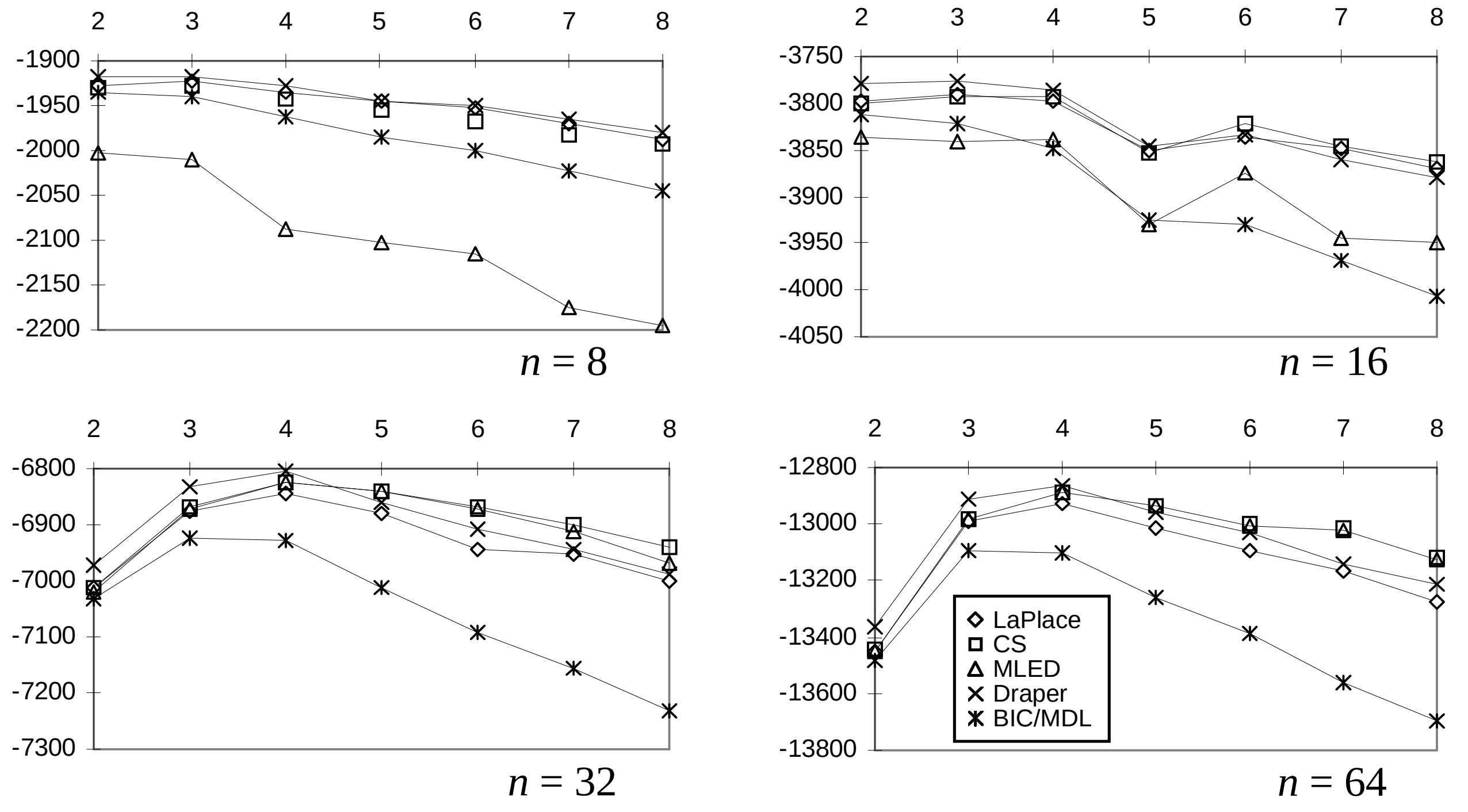

Figure 1: Approximate log marginal likelihood of the data given a test model as a function of the number of hidden states in the test model. The 400 sample data sets were generated from naive-Bayes models with $n$ observed variables and 4 hidden states.

and ran one E and M step. Then, we retained the 32 copies of the parameters for which $g(\boldsymbol{\theta}_s)$ was largest, and ran two EM iterations. Next, we retained the 16 copies of the parameters for which $g(\boldsymbol{\theta}_s)$ was largest, and ran 4 EM iterations. We continued this procedure four more times, until only one set of parameters remained.

To guarantee convergence of the EM algorithm, we performed 200 EM iterations following the initialization phase. To check that the algorithm had converged, we measured the relative change of $g(\boldsymbol{\theta}_s)$ between successive iterations. Using a convergence criterion similar to that of AutoClass' default, we said that the EM algorithm had converged when this relative change fell below 0.00001. The algorithm converged in all but one of the 550 runs.

We assigned Dirichlet priors to each parameter set $\boldsymbol{\theta}_{ij}$. We used the almost uniform prior $\alpha_{ijk} = 1+\epsilon$, because it produced local maxima in the interior of the parameter space. (The traditional Laplace approximation is not valid at the boundary of a parameter space.) Our conclusions are not sensitive to $\epsilon$ in the range we tested (0.1 to 0.001). We report results for the value $\epsilon = 0.01$.

As described by Equation 8, we evaluated the Laplace approximation at the MAP of $\boldsymbol{\theta}_s$. To simplify the computations, we also evaluated the CS, MLED, Draper, and BIC/MDL measures at the MAP. Given our choice for $\alpha_{ijk}$, the differences between the MAP and ML values were insignificant. We used the method of Thiesson (1995) to evaluate the negative Hessian of $g(\boldsymbol{\theta}_s)$. To compute the CS scoring function, we assumed that dimensions $d'$ and $d$ are equal. Although we have no proof of this assumption, experiments in Geiger et al. (in this proceedings) suggest that the assumption is valid.

All experiments were run on a P5 100MHz machine under the Windows NT$^{\text{TM}}$ operating system. The algorithms were implemented in C++.

## 4 Results and Discussion

First, we evaluated the approximations for use in model averaging, comparing plots of approximate log marginal likelihood versus the number of states of the hidden variable in the test model. We conducted three sets of comparisons for different values of $c$ (number of states of the hidden variable), $n$ (number of observed variables), and $N$ (sample size). The results are almost the same for different data sets in a given experiment (if we were to show one-standard deviation errors bars, they would be invisible for most data points, and barely visible for the remaining points). Consequently, we show results for only one data set

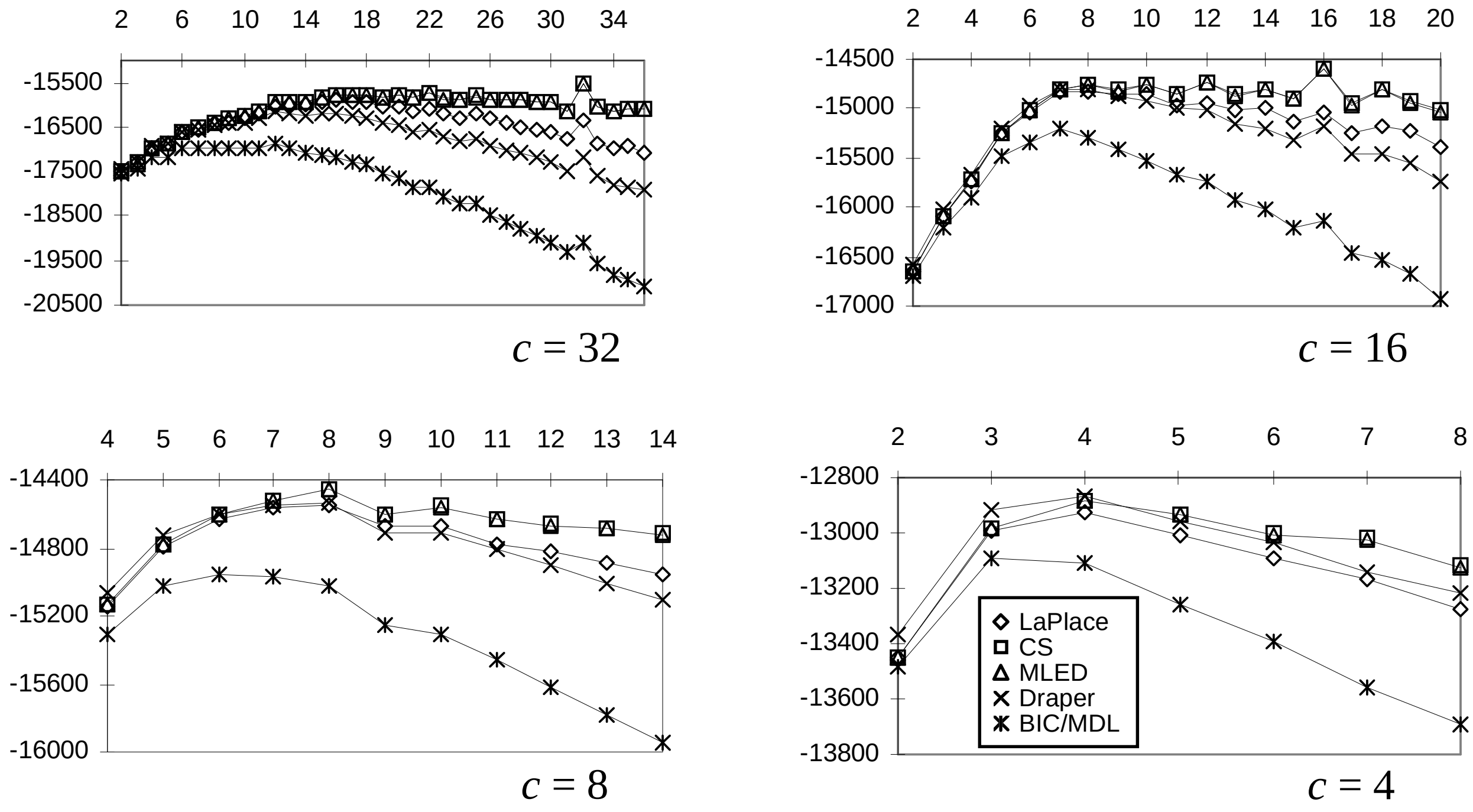

Figure 2: Approximate log marginal likelihood of the data given a test model as a function of the number of hidden states in the test model. The 400 sample data sets were generated from naive-Bayes models with 64 observed variables and $c$ hidden states.

per experiment.

In our first set of experiments, we fixed $c = 4$ and $N = 400$ and varied $n$. In particular, we generated 400-sample data sets from four naive-Bayes models with 8, 16, 32, and 64 observed variables, respectively, each model having a hidden variable with four states. Figure 1 shows the approximate log marginal likelihood of a data given test models having hidden variables with two to eight states. (Recall that each test model has the same number of observed variables as the corresponding generative model.)

In our second set of experiments, we fixed $n = 64$ and $N = 400$, and varied $c$. In particular, we generated 400-sample data sets from four naive-Bayes models with $c = 32$, 16, 8, and 4 hidden states respectively, each model having 64 observed variables. Figure 2 shows the approximate log marginal likelihood of a data for test models having values of $c$ that straddle the value of $c$ for the generative model.

In our third set of experiments, we fixed $n = 32$ and $c = 4$, and varied $N$. In particular, from a naive-Bayes model with $n = 32$ and $c = 4$, we generated data sets with sample sizes ($N$) 100, 200, 400, and 800, respectively. Figure 3 shows the approximate log marginal likelihood of a data for test models having hidden variables with two to eight states.

The trends in the marginal-likelihood curves as a function of $n$, $c$, and $N$ are not surprising. For each approximation, the curves become more peaked about the value of $c$ (the number of hidden states in the generative model) as (1) $N$ increases, (2) $n$ increases, and (3) as $c$ decreases. The first result says that learning improves as the amount of data increases. The second result is a reflection of the fact that larger numbers of observed variables provide more evidence for the identity of the hidden variable. The third result says that it becomes more difficult to learn as the number of hidden states increases.

In comparing these curves, note that only differences in the shape of the curves are important. The height of the curves are not important, because the marginal likelihoods (i.e., relative posterior probabilities) are normalized in the process of model averaging. Overall, the Draper and CS scoring functions appear to be equally good approximations, both better than the BIC/MDL. The MLED and CS scoring functions are almost identical, except for small values of $n$, where the CS approximation is better.

Next, we evaluated the approximations for use in model selection. In each experiment for a particular $n$, $c$, and $N$, we computed the size of the model selected by a given approximation—that is, the number of states of the hidden variable in the test model hav-ing the largest approximate marginal likelihood. We then subtracted from this number the size of the model selected by the gold-standard Laplace approximation, yielding a quantity called $\Delta c$. The results are shown in Table 1.

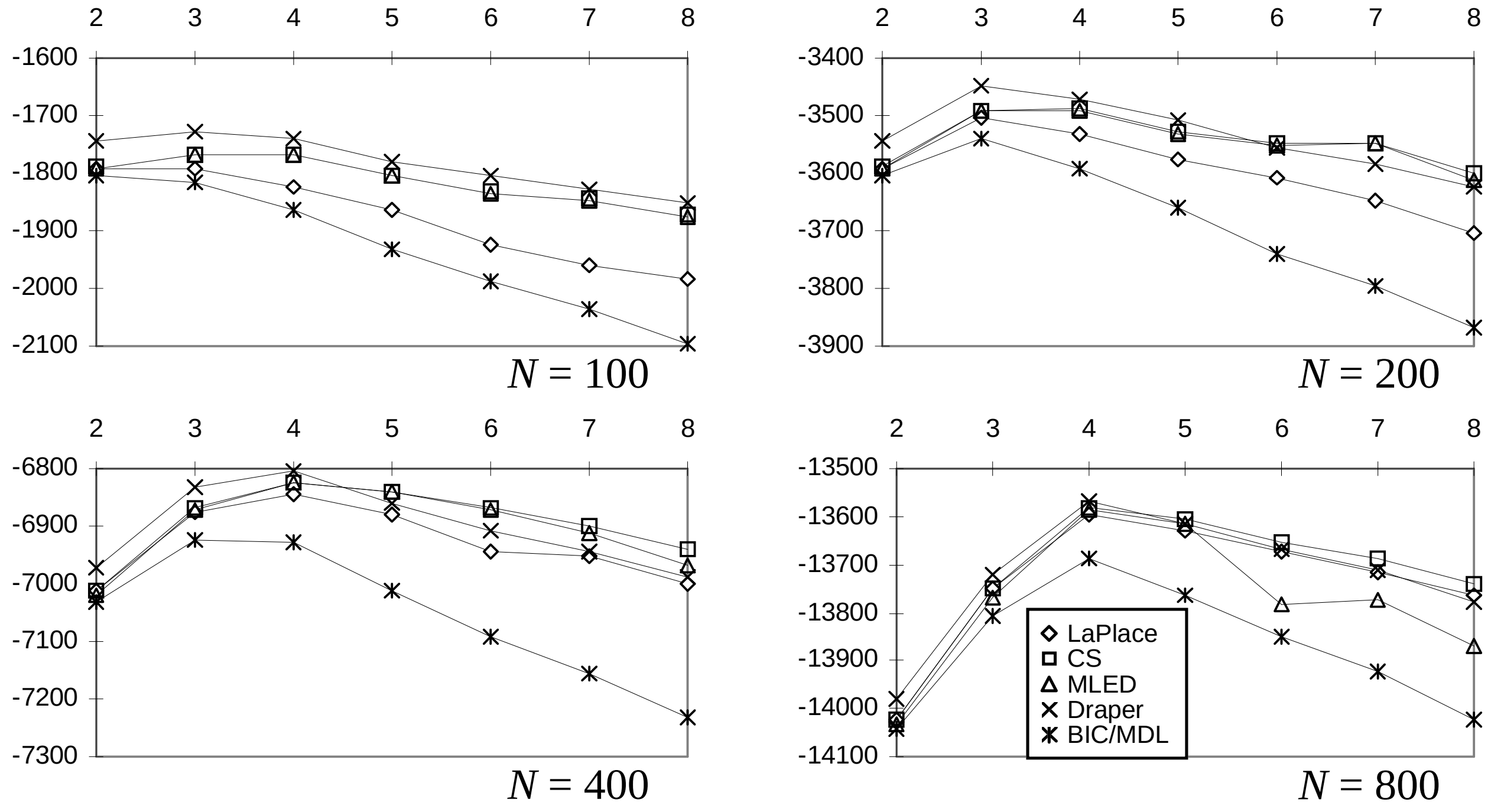

Figure 3: Approximate log marginal likelihood of the data given a test model as a function of the number of hidden states in the test model. The data sets of size $N$ were generated from naive-Bayes models with 32 observed variables and 4 hidden states.

Overall, the Draper, CS, and MLED scoring functions are more accurate than the BIC/MDL, which consistently selects models that are too simple. The CS and MLED results are almost the same, except for small values of $n$ where CS does slightly better. The CS and MLED measures tend to select models that are too complex ($\Delta c > 0$), whereas the Draper measure tends to select models that are too simple ($\Delta c < 0$).

For large values of $c$, the Draper measure does better than the CS approximation for large values of $c$. The source of this difference can be seen in the top two graphs of Figure 2. All algorithms in both these graphs show two peaks: a broad peak around $c/2$ and a sharp peak around $c$, where $c$ is the number of hidden states in the generative model. The Laplace and Draper (and BIC) curves tend downward on the right sharply enough such that the first peak dominates. In contrast, the CS curve remains fairly flat to the right such that the second peak dominates. The existence of the second peak at the true number of hidden states is quite interesting (and unexpected), but we do not pursue it further in this paper.

As we have discussed, the accuracy results must be balanced against the computational costs of the various approximations. The time complexities given in Section 2 are overly pessimistic for naive-Bayes models, because probabilistic inferences can be cached and reused. For naive-Bayes models, the evaluation of the CS, MLED, Draper, and BIC/MDL measures (again dominated by the MAP computation) has time complexity $O(edN)$. The evaluation of Laplace approximation (again dominated by the determination of the Hessian), is given by $O(d^2N)$. To appreciate the constants for these costs, the run times for the experiment $n = 64, c = 32, N = 400$ are shown in Table 2.

The bottom line is that the Draper and CS measures are both accurate and efficient, and are probably the approximations of choice for most applications.

Our findings are valid only for naive-Bayes models with a hidden root node. These results are important, because they apply directly to the AutoClass algorithm, which is growing in popularity. Also, it is likely that our results will extend to models for discrete variables and data sets where each variable that is unobserved has an observed Markov blanket. Under these conditions, each Bayesian inference required by the scoring functions (e.g., Equation 11) reduces to a naive-Bayes computation. Nonetheless, more detailed experiments are warranted to address models with more general structure and non-discrete distributions. Finally, we again note that our results do not rule out the possibility that the CS or MLED approximations are better than the Laplace approximation.

Table 1: Errors in model selection (mean ± s.d.).

| Experiment | | | $\Delta c$ | | | |
|---|---|---|---|---|---|---|
| $n$ | $c$ | $N$ | CS | MLED | Draper | BIC/MDL |
| 8 | 4 | 400 | 0 | 0.4 ± 1.5 | 0 | -0.2±0.4 |
| 16 | 4 | 400 | 0.2 ± 0.4 | -0.2 ± 0.8 | 0.2 ± 0.4 | -0.8±0.4 |
| 32 | 4 | 400 | 0 | 0 | 0 | -0.4±0.5 |
| 64 | 4 | 400 | 0 | 0 | 0 | -0.2±0.4 |
| 64 | 32 | 400 | 16.2 ±1.5 | 16.2 ± 1.5 | -2.2 ± 2.0 | -6.0±2.7 |
| 64 | 16 | 400 | 5.0 ±6.4 | 5.0 ± 6.4 | -1.6 ± 1.1 | -3.0±1.4 |
| 64 | 8 | 400 | 0.8 ±0.8 | 0.8 ± 0.8 | 0 | -1.0±1.0 |
| 64 | 4 | 400 | 0 | 0 | 0 | -0.2±0.4 |
| 32 | 4 | 100 | 0.6 ± 0.9 | 0.6 ± 0.9 | 0 | -0.6±0.5 |
| 32 | 4 | 200 | 0.2 ± 0.4 | 0.2 ± 0.4 | 0 | -0.6±0.5 |
| 32 | 4 | 400 | 0 | 0 | 0 | -0.4±0.5 |
| 32 | 4 | 800 | 0 | 0 | 0 | 0 |

## 5 Reality Check

In our analysis of scoring functions for hidden-variable models, we have made an important assumption. Namely, we have assumed that, when the true model contains a hidden variable, it is better to learn by searching over models with hidden variables than those without hidden variables. This assumption is not trivially correct. Given a naive-Bayes model for the variables $C, X_1, \ldots, X_n$, the joint distribution for these variables can be encoded by a Bayesian network without hidden variables. (Assuming there are no accidental cancellations in the probabilities, this Bayesian network will be completely connected.) Thus, we can attempt to learn a model containing no hidden variables, and this model may be more accurate than that learned by searching over naive-Bayes models having a hidden root node.

We tested our assumption as follows. First, we generated a naive-Bayes model with $n = 12$ and $c = 3$. From this model we generated a data set of size 800, discarding the observations of the variable $C$. Second, we learned a single naive-Bayes model containing a hidden root node using the experimental technique described in the previous section. In particular, we varied the number of hidden states of the naive-Bayes model, and selected the one with the largest (approximate) marginal likelihood. (In this case, all scoring functions yielded the same model: one with three hidden states). Third, we learned a single model containing no hidden variables using the approach described in Heckerman et al. (1995). In particular, we used the BD scoring function with a uniform prior over the parameters in conjunction with a greedy search algorithm (in directed-graph space) initialized with an empty graph.

We evaluated the two learned models by comparing their marginal likelihoods. Specifically, we computed $\Delta m \equiv \log p(D|S^h_{\text{hidden}}) - \log p(D|S^h_{\text{nohide}})$. We used the Laplace approximation to compute the first term, and the exact expression for marginal likelihood (e.g., Heckerman et al., 1995) to compute the second term. Repeating this experiment five times, we obtained $\Delta m = 26 \pm 33$, indicating that the hidden-variable model better predicted the data. In additional experiments, we found that $\Delta m$ increased as we increased the size of the models.

## 6 Conclusions

We have evaluated the Laplace, CS, MLED, Draper, and BIC/MDL approximations for the marginal likelihood of naive-Bayes models with a hidden root node, under the assumption that the Laplace approximation is the most accurate of the scoring functions. Our experiments indicate that (1) the BIC/MDL measure is the least accurate, having a bias in favor of simple models, and (2) the Draper and CS measures are the most accurate, having a bias in favor of simple and complex models, respectively, in most cases.

## Acknowledgments

We thank Dan Geiger and Chris Meek for useful discussions about asymptotic approximations, Koos Rommelse for his help with system implementation, and the anonymous reviewers for their suggestions.

Table 2: Algorithm run times (in seconds) as a function of the dimension of the test model for the experiment $n = 64, c = 32$, and $N = 400$. The values for EM are times to convergence. The values for the scoring functions exclude EM run times.

| $d$ | EM | Laplace | CS | MLED | Draper | BIC/MDL |
|---|---|---|---|---|---|---|
| 1689 | 500 | 2800 | 2 | 0.06 | 2 | 2 |
| 1754 | 580 | 3000 | 2 | 0.07 | 2 | 2 |
| 1819 | 590 | 3300 | 2 | 0.07 | 2 | 2 |
| 1884 | 650 | 3600 | 2 | 0.08 | 2 | 2 |
| 1949 | 680 | 3900 | 2 | 0.08 | 2 | 2 |
| 2014 | 680 | 4200 | 2 | 0.08 | 2 | 2 |
| 2079 | 530 | 4600 | 2 | 0.08 | 2 | 2 |
| 2144 | 760 | 4900 | 2 | 0.08 | 2 | 2 |
| 2209 | 810 | 5500 | 2 | 0.08 | 2 | 2 |
| 2274 | 790 | 5900 | 3 | 0.08 | 3 | 3 |
| 2339 | 870 | 6700 | 3 | 0.08 | 3 | 3 |